\documentclass[letterpaper]{article} 
\usepackage[submission]{aaai23} 
\usepackage[pdftex]{graphicx}
\usepackage{times}  
\usepackage{helvet}  
\usepackage{courier}  
\usepackage[hyphens]{url}  
\usepackage{natbib}  
\usepackage{caption} 
\usepackage{algorithm}
\usepackage{algorithmic}

\usepackage{amsmath,bm}
\usepackage{amsfonts} 
\usepackage{xcolor}
\usepackage{subcaption}
\usepackage[normalem]{ulem} 

\usepackage{newfloat}
\usepackage{listings}
\DeclareCaptionStyle{ruled}{labelfont=normalfont,labelsep=colon,strut=off} 
\floatstyle{ruled}
\newfloat{listing}{tb}{lst}{}
\floatname{listing}{Listing}
\title{Fundamentals of Task-Agnostic Data Valuation}
\author{
    Mohammad Mohammadi Amiri\textsuperscript{\rm 1},
    Fr\'ed\'eric Berdoz\textsuperscript{\rm 2},
    Ramesh Raskar\textsuperscript{\rm 1}
}
\affiliations{
    \textsuperscript{\rm 1}MIT Media Lab, email:\{mamiri, raskar\}@mit.edu\\
    \textsuperscript{\rm 2}EPFL, email: frederic.berdoz@epfl.ch

}

\usepackage{bibentry}

\begin{document}

\maketitle

\begin{abstract}
We study valuing the data of a data owner/seller for a data seeker/buyer.
Data valuation is often carried out for a specific task assuming a particular utility metric, such as test accuracy on a validation set, that may not exist in practice. 
In this work, we focus on task-agnostic data valuation without any validation requirements. 
The data buyer has access to a limited amount of data (which could be publicly available) and seeks more data samples from a data seller. 
We formulate the problem as estimating the differences in the statistical properties of the data at the seller with respect to the baseline data available at the buyer. 
We capture these statistical differences through second moment by measuring \textit{diversity} and \textit{relevance} of the seller's data for the buyer; 
we estimate these measures through queries to the seller without requesting raw data. 
We design the queries with the proposed approach so that the seller is blind to the buyer's raw data and has no knowledge to fabricate responses to queries to obtain a desired outcome of the diversity and relevance trade-off.
We will show through extensive experiments on real tabular and image datasets that the proposed estimates capture the diversity and relevance of the seller's data for the buyer.  
\end{abstract}
 
\vspace{-.25cm} 
\section{Introduction}
Data is the main fuel of the modern world enabling artificial intelligence and driving innovation and technological growth.
The demand for data has grown substantially, and it is extremely valuable for sectors to acquire high quality data to discover knowledge and improve their products and services. 
As the demands for data have grown substantially, data products have become valuable assets to purchase and sale.
This calls for establishing a data marketplace that connects different parties and facilitates trading data.

A data marketplace mainly includes three components, data sellers, broker, and data buyers; data sellers own the data and share it with the broker in exchange for rewards; data buyers want to acquire data, and broker facilitates trading data. 
As a valuable resource, it is important to establish a principled method to quantify the worth of the sellers' data and its value for the buyers. 
This is addressed via data valuation which is the essential component for realization of a fair marketplace for sellers and buyers.
Data valuation arises in various applications such as collaborative machine learning (\citeauthor{CMLIncentiveAware,CMLIncentivizeCollabML}), federated learning (\citeauthor{FLProfitAllocation,FLBoudgetBiynededIncentives}), data marketing (\citeauthor{SurveyMarketplacesForData,PricingApproachesForFataMarkets}), advertisement (\citeauthor{DataCompetitionDigitalPlatforms,OptimalAdvertisingInformationProducts}), recommendation systems (\citeauthor{RecommIncentivizingExploration,RecommenderSystemsMechanisms}), and data sharing (\citeauthor{DataSharingMarkets,ParetoDataSharing}).

Data valuation is carried out either based on ``intrinsic'' or ``extrinsic'' factors;
intrinsic data valuation is data-driven and based on the quality of dataset (\citeauthor{UnlockingValuePrivacyTrading,DataSupportAIAllPricingValuationGovernance}), while extrinsic data valuation considers demand-supply and game-theoretic mechanisms (\citeauthor{DataCollectionWirelessCommunicationIoT,SurveyDataPricingMethods}).
It is a common practice to couple intrinsic data valuation with a utility metric for validation (\citeauthor{DataShapleyEquitableValuation,EfficientValuationBasedShapley}), or with a specific machine learning (ML) task (\citeauthor{MarketplaceDataAlgorithmicSolution,ModelBasedPricingMarketplace}). 
In particular, for ML applications, data is often valued assuming existence of a validation set using validation accuracy as a metric (\citeauthor{PrincipledApproachDataValuationForFL,YouLoveTheCore}).
Also, ML models trained with a target task are used to estimate the value of the data used for training the models (\citeauthor{SurveyDataPricingEconomicsDataScience,DEALEREnd2EndModelMarketplaceDP}). 
On the other hand, extrinsic data valuation techniques consider external factors such as competition and demands (\citeauthor{TowardsDataAuctionsExternalities,InformationSaleCompetition}), which requires estimating costumers' demands for products and competitors price levels to price a product (\citeauthor{PricingStrategiesCorporateProfitability,DataPricingMachineLearningPipelines}).

Enforcing a close coupling between intrinsic data valuation and existence of a validation set may not be practical since a validation set that all the parties agree on may not exist, and a particular validation set may not sufficiently represent the data distribution for a learning task (\citeauthor{ValidationFreeReplicationRobustVolume}).
Furthermore, having a validation set may provide the chance to malicious sellers to modify their datasets to overfit on the validation set. 
Also, considering a specific ML model/task for data valuation may not be aligned with the interests of all the parties.   
We instead take a step back and consider an intrinsic data valuation without any validation requirements and before performing any tasks such as training a ML model. 
We take a step towards addressing the challenge of formulating a model- and task-agnostic intrinsic valuation of data at a seller for a buyer.
The authors in (\citeauthor{ValidationFreeReplicationRobustVolume}) develop a technique independent of validation based solely on the diversity of seller's data, which captures the variation/dissimilarity across data samples; 
this provides the same value of data at a seller for all the buyers.
However, we believe that diversity of data alone may not be sufficient for data valuation for two reasons. 
First, performing data valuation independent of the buyers makes it hard to realize the relevance of the seller's data for the buyer.
Consider the case when a buyer is interested in health data, such as chest X-ray images, while a seller has a very diverse set of images of animals. 
Thus, the inherently diverse dataset at the seller is irrelevant for the buyer, and this needs to be captured by data valuation.
Additionally, a seller can fabricate data to increase its diversity through, for example, adding random noise.

We focus on an intrinsic task-agnostic data valuation considering the fact that \uline{data at each seller has a distinct value for each buyer} (\citeauthor{DataSupportAIAllPricingValuationGovernance}).
We measure the value of data at a seller in dependence with the already available data at the buyer, some of which may be publicly available, which stays local at the buyer and is not shared with any parties.
This provides a unique valuation of a seller's data for each buyer. 
We aim to value the data through comparing the statistical properties of the two datasets and formulate the problem as estimating \textbf{diversity} and \textbf{relevance} of the seller's data for the buyer.
We then estimate diversity and relevance by measuring the differences and similarities in the statistical properties of the two datasets through second moment. 
This is carried out through queries from the buyer to the seller designed such that it is infeasible for the seller to fabricate responses to the queries and manipulate the data to achieve a desired outcome of diversity and relevance pair.

\textit{Notations}: 
A multi-variate normal distribution with mean vector $\boldsymbol{\mu}$ and covariance matrix $\boldsymbol{\Sigma}$ is denoted by $\mathcal{N} (\boldsymbol{\mu}, \boldsymbol{\Sigma})$; $\boldsymbol{0}_{n}$ represents an all-zero vector of dimension $n$; $l^2$ norm of a vector is denoted by $\| \cdot \|$; $\mbox{Diag} (\lambda_1, ..., \lambda_d)$ returns a $d \times d$ diagonal matrix with diagonal entries $\lambda_1, ..., \lambda_d$. 
Cardinality of a set is shown by $|\cdot|$.


\vspace{-.0cm}
\section{Problem Motivation and Formulation}\label{Sec_ProbFor}
We consider a data marketplace with an arbitrary number of buyers and sellers, assuming that each buyer (she) has access to some data samples\footnote{This could be private data at each buyer or a publicly available dataset or a combination of both.} and wants to buy extra data from one or multiple sellers. 
The goal is to measure the value of the data at a seller (he) for a buyer without focusing on a specific task for which the buyer is buying data.
Data as a random variable is entirely defined by its distribution, and data distribution contains all the statistical information about the data.
As a result, comparing data distributions at the buyer and seller could provide a comprehensive means for data valuation.
However, in practice (as for ML applications), the data distribution is unknown, and it is often computationally impossible to approximate it using only a limited number of samples.
Hence, we may instead directly use the data samples as the realizations of their distribution to capture their statistical properties. 
We further argue that the differences and similarities in the statistical properties of the data at different parties are reflected by two metrics, \textbf{diversity} and \textbf{relevance}.
Accordingly, we aim to estimate these two metrics using the data at the seller and the buyer for data valuation.

Diversity measures how much of different statistical properties the seller's data adds to the buyer's data, where we note that her data is limited to capture all the statistical properties of the original distribution.
Whereas, relevance captures the similarity in the statistical properties of the two datasets. 
Consider a buyer with some cat and dog images, both with only black color.
Intuitively, images of colorful cats and dogs seem to be a perfect addition to the buyer's data, where it provides some statistical properties that the buyer's data has not seen (because of the difference in colors), and some similarity in the statistical properties (having the same animals).
Other images (except cats and dogs) could provide highly diverse data for the buyer; however, the relevance may be very limited, which prevents the buyer's data to capture the entire distribution.
On the contrary, a dataset with black cats and dogs is highly relevant to the buyer's data, while it may not add any new statistics to it. 

Another example is that of sample complexity in ML, which, given a data distribution, is defined as the minimum number of independent and identically distributed samples required for the ML model to generalize to that distribution without overfitting.
Adding diverse data to the buyer's data helps the ML model to cover a wider range of statistical properties; 
however, this will require a larger sample size to guarantee that the model can generalize well to the new (statistically more diverse) dataset. 
While, after receiving a more relevant dataset at the buyer, it is likely that the ML model generalizes well (satisfy the sample complexity requirements) to the (limited) data statistics.
As a result, there is a trade-off between the amount of diversity and relevance that the buyer is willing to receive and the performance (in this case whether the model generalizes well to capture all the statistical properties of the data).

\begin{figure*}[t!]
\centering
\begin{subfigure}{.245\textwidth}
  \centering
  \includegraphics[width=1\linewidth]{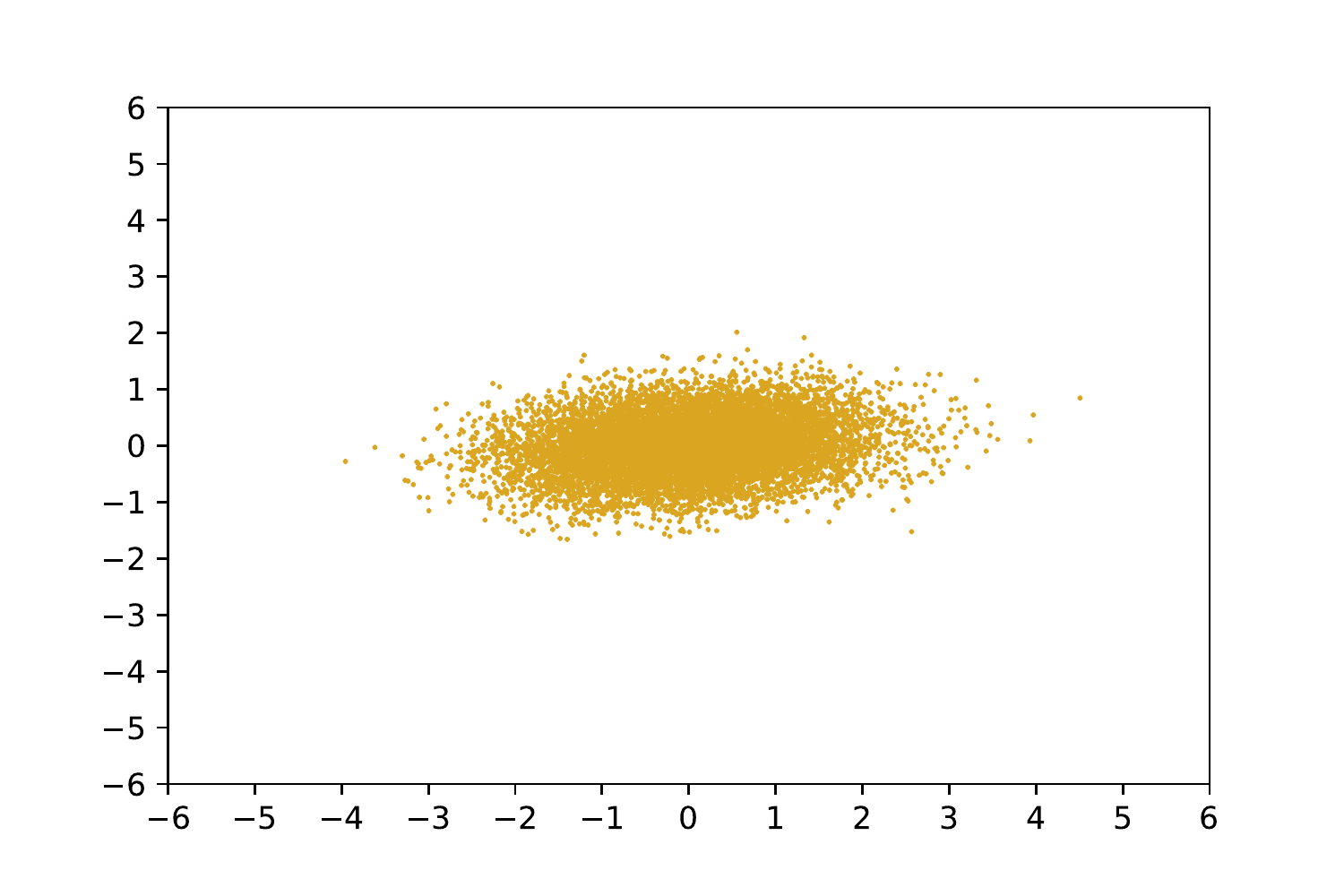}
  \caption{Buyer}
  \label{Buyer_scatters}
\end{subfigure}%
\begin{subfigure}{.245\textwidth}
  \centering
  \includegraphics[width=1\linewidth]{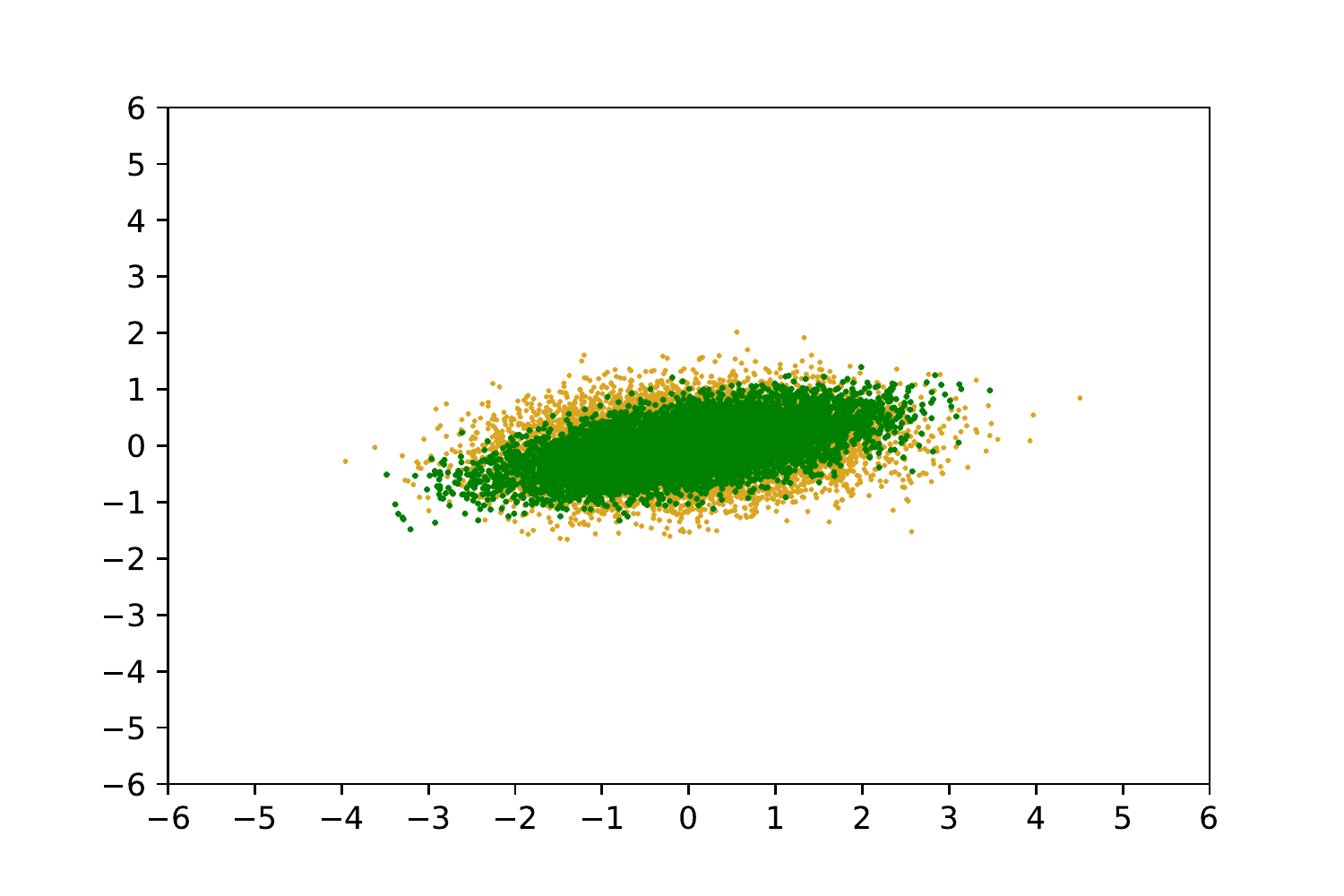}
  \caption{Seller 1 vs. buyer}
  \label{Seller1_scatters}
\end{subfigure}
\begin{subfigure}{.245\textwidth}
  \centering
  \includegraphics[width=1\linewidth]{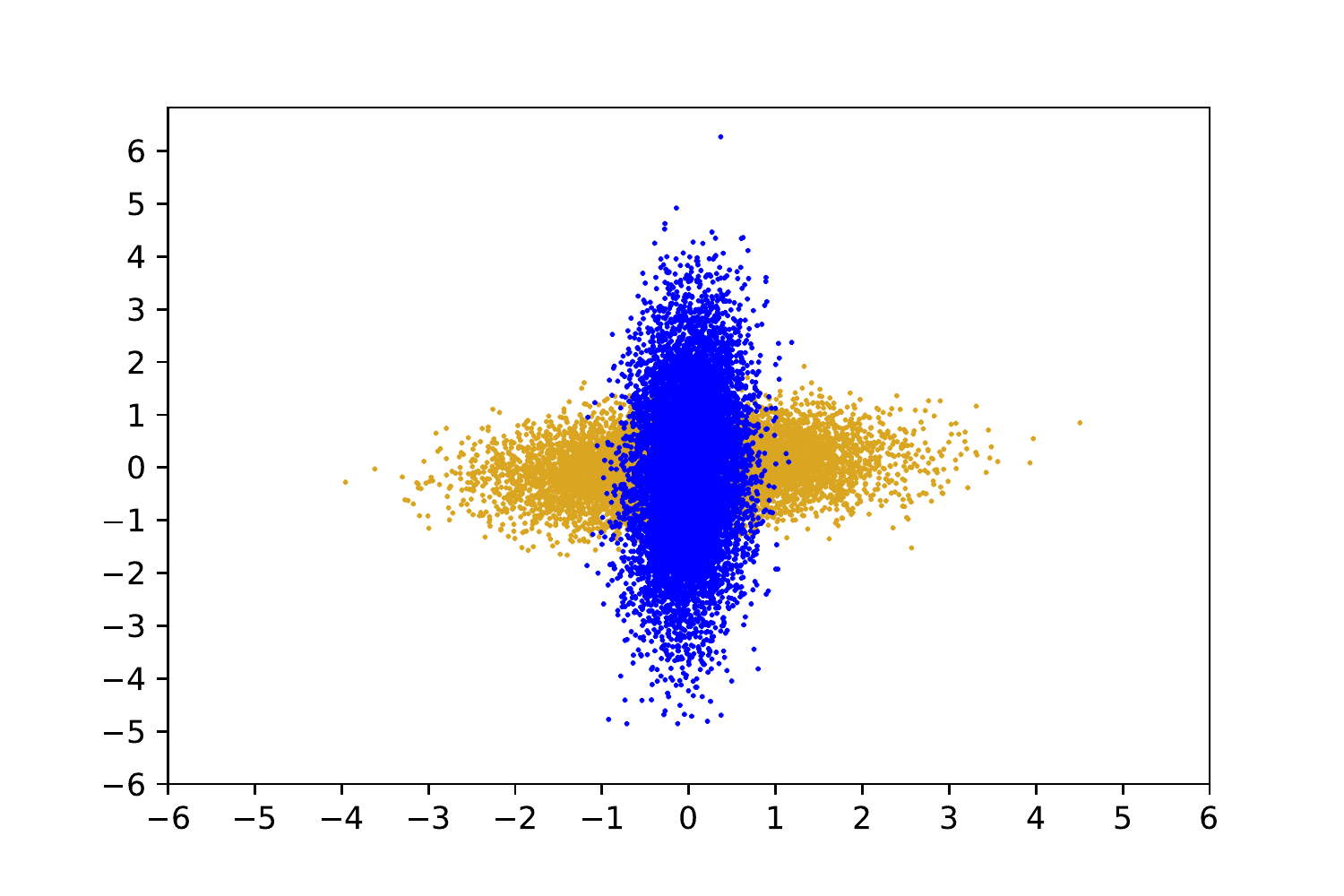}
  \caption{Seller 2 vs. buyer}
  \label{Seller2_scatters}
\end{subfigure}
\begin{subfigure}{.245\textwidth}
  \centering
  \includegraphics[width=1\linewidth]{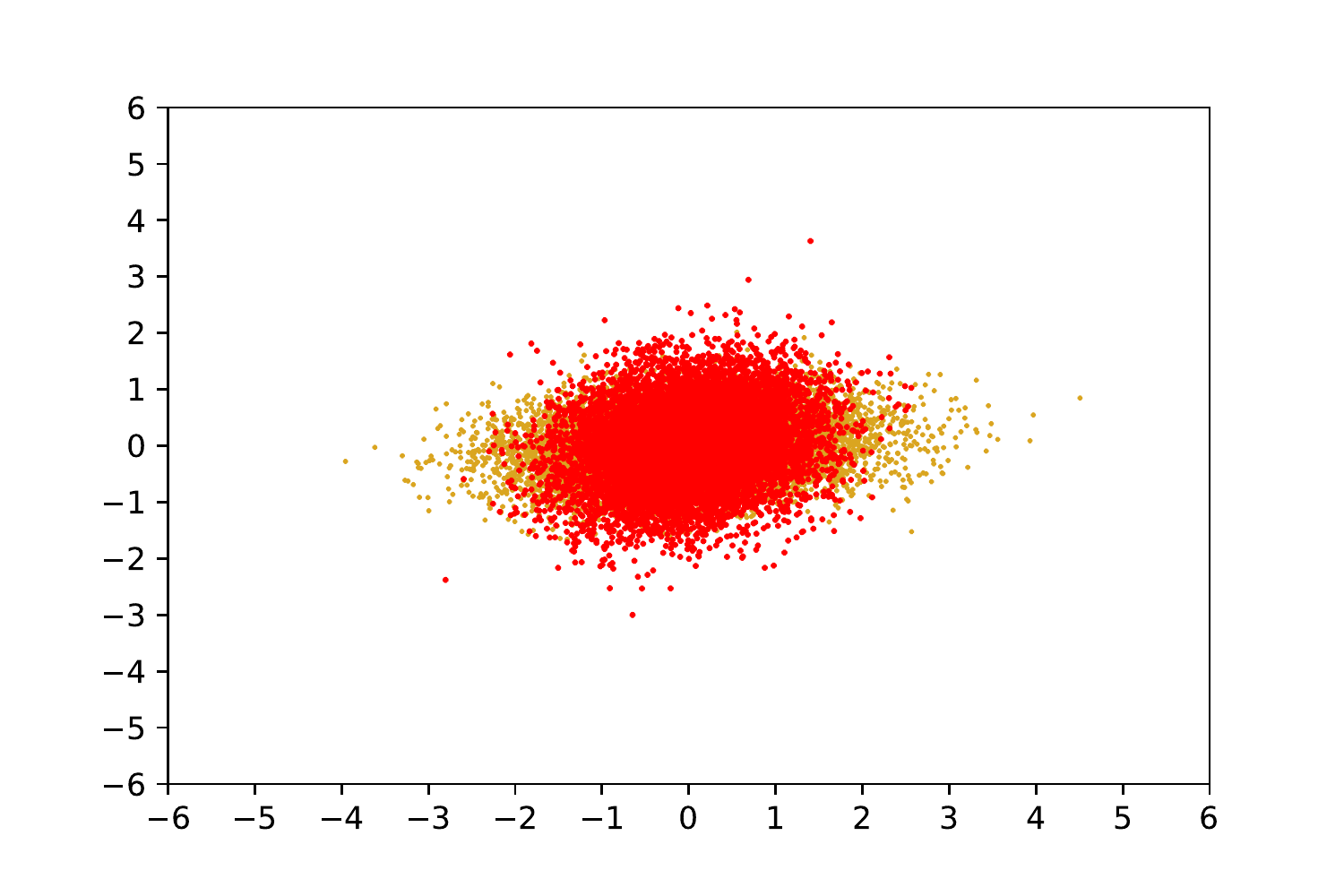}
  \caption{Seller 3 vs. buyer}
  \label{Seller3_scatters}
\end{subfigure}
\caption{Data scatters illustration in 2-D for buyer and sellers 1 to 3's data with covariance matrices $[[1, 0.1], [0.1, 0.25]]$, $[[0.9, 0.2], [0.2, 0.15]]$, $[[0.1, 0.05], [0.05, 2]]$, and $[[0.5, 0.1], [0.1, 0.5]]$, respectively.}
\label{All_scatters}
\end{figure*}

Our goal is to develop a task-agnostic data valuation through measuring diversity and relevance between two datasets.
We consider buyer's data as the baseline dataset and measure the diversity and relevance of a seller's data with respect to this baseline.
Let us denote the buyer's and seller's data with matrices $\boldsymbol{B} \in \mathbb{R}^{n_b \times d}$ and $\boldsymbol{S} \in \mathbb{R}^{n_s \times d}$, respectively.
The underlying assumption is that the datasets at the buyer and seller have the same feature space and as in ML applications have been zero-centered and normalized (this will guarantee that the datasets have the same support set).
Data valuation is defined assuming that data could be readily used at the buyer without any computationally heavy post-processing (except zero-centering and normalization).

Considering the buyer's dataset $\boldsymbol{B}$ as the baseline, we denote the diversity and relevance of another dataset with respect to the baseline dataset by $D_{\boldsymbol{B}}$ and $R_{\boldsymbol{B}}$, respectively, such that $D_{\boldsymbol{B}}:\mathbb{R}^{n_s \times d} \to [0, 1]$ and $R_{\boldsymbol{B}}:\mathbb{R}^{n_s \times d} \to [0, 1]$.
Accordingly, both diversity and relevance accept a dataset $\boldsymbol{S}$ (seller's data matrix) as input and map it to a real number in the interval $[0, 1]$. 
According to the above definition, the output of $D_{\boldsymbol{B}}$ and $R_{\boldsymbol{B}}$ is general enough, since any bounded interval can be normalized to the interval $[0, 1]$.
A larger $D_{\boldsymbol{B}}$ ($R_{\boldsymbol{B}}$) indicates a larger diversity (relevance) of a dataset with respect to $\boldsymbol{B}$.
As a result, for any specific realization of the measures $D_{\boldsymbol{B}}$ and $R_{\boldsymbol{B}}$ defined above, being close to $0$ indicates the minimum diversity (relevance), while a measure close to $1$ translates into the maximum diversity (relevance) of a dataset compared to the baseline dataset $\boldsymbol{B}$.

Any realization of $D_{\boldsymbol{B}}$ and $R_{\boldsymbol{B}}$ should satisfy the following two intuitive cases:
\begin{itemize}
\item \textbf{Case 1:} $D_{\boldsymbol{B}} (\boldsymbol{B})=0$; $R_{\boldsymbol{B}} (\boldsymbol{B}) = 1$; that is, the same dataset has no diversity and maximum relevance.
\item \textbf{Case 2:} Using any distance measure, if the distance between the distributions of $\boldsymbol{B} \in \mathbb{R}^{n_b \times d}$ and $\boldsymbol{S} \in \mathbb{R}^{n_s \times d}$ is unbounded\footnote{This is for any reasonable distance metric between two datasets, for instance Kullback–Leibler divergence, or R\'enyi divergence. For distance metrics with upper bound, such as Jensen–Shannon divergence, this could be rewritten as the maximum distance between the two datasets.}, we have $D_{\boldsymbol{B}} (\boldsymbol{S})=1$; $R_{\boldsymbol{B}} (\boldsymbol{S}) = 0$; that is, for distributions $P_b$ and $P_s$ of data at the buyer and seller, respectively, with their distance denoted by $\ell(P_b, P_s)$, $\lim_{\ell(P_b, P_s)\to \infty} D_{\boldsymbol{B}} (\boldsymbol{S})=1$ and $\lim_{\ell(P_b, P_s)\to \infty} R_{\boldsymbol{B}} (\boldsymbol{S})=0$. 
\end{itemize}

We highlight that, unlike the above task-agnostic data valuation formulation, a task-dependent data valuation may be a function of a learning algorithm, which takes as input a training dataset and outputs a ML model;
furthermore, it may depend on a utility function which takes as input the output of the learning algorithm (ML model) and/or a dataset and outputs a real value performance score (\citeauthor{DVIngredientsStrategiesOpenChallenges}).
Next we motivate our approach to measure diversity and relevance through a simple example.

\vspace{-.0cm}
\section{Motivating Example}

Here we focus on a 2-D feature space, i.e., $d=2$, where $\boldsymbol{B} \in \mathbb{R}^{n_b \times 2}$ and $\boldsymbol{S} \in \mathbb{R}^{n_s \times 2}$. 
We consider the case where entries of data matrices $\boldsymbol{B}$ and $\boldsymbol{S}$ are distributed according to $\mathcal{N} (\boldsymbol{0}_{2}, \boldsymbol{\Sigma}_b)$ and $\mathcal{N} (\boldsymbol{0}_{2}, \boldsymbol{\Sigma}_s)$, respectively, where we note that data distribution is unknown to the nodes.
For simplicity, we assume that the number of data samples at the buyer and each seller is $10^4$, i.e., $n_b = n_s = 10^4$.
We aim to measure the diversity and relevance of various datasets with respect to the baseline dataset (buyer's data) with a covariance matrix $\boldsymbol{\Sigma}_b = [[1, 0.1], [0.1, 0.25]]$.
Fig. \ref{Buyer_scatters} illustrates the scatters of the buyer's data in 2-D. 
We observe that the buyer's data is scattered mostly across the first dimension.

We consider five sellers with various datasets.
The datasets in the first three sellers have covariance matrices $\boldsymbol{\Sigma}_{s_1} = [[0.9, 0.2], [0.2, 0.15]]$, $\boldsymbol{\Sigma}_{s_2} = [[0.1, 0.05], [0.05, 2]]$, and $\boldsymbol{\Sigma}_{s_3} = [[0.5, 0.1], [0.1, 0.5]]$, respectively.
Figs. \ref{Seller1_scatters}, \ref{Seller2_scatters}, and \ref{Seller3_scatters} demonstrate the scatters of the first three sellers' data compared to the buyer's data in 2-D.
According to the covariance matrices and Fig. \ref{All_scatters}, it is intuitive to conclude that, having the buyer's data as baseline, seller 1's data is the most similar compared to the data of the other two sellers, while seller 2's data has the least similarity among the three sellers.
We expect seller 3's data to have some level of similarity and some level of difference compared to the buyer's data.
We further consider sellers 4 and 5 with datasets with covariance matrices $\boldsymbol{\Sigma}_{s_4} = [[1, 0.1], [0.1, 0.25]]$ and $\boldsymbol{\Sigma}_{s_5} = [[50, 0], [0, 50]]$, respectively.
It is expected that seller 4's data with the same covariance matrix as the buyer's data should result in Case 1, i.e., minimum diversity and maximum relevance, while seller 5 with extremely different data distribution compared to the buyer's data should lead to Case 2, i.e., maximum diversity and minimum relevance.

\begin{figure}[t!]
\centering
\includegraphics[scale=.91]{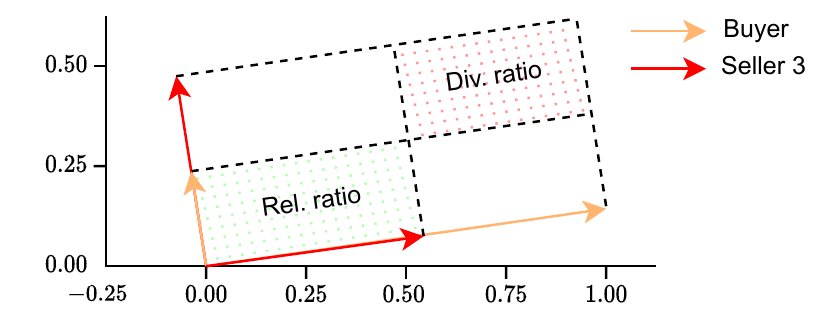}
\caption{Variance of the buyer's and seller 3's data in the directions of the principal components of buyer's data.}
\label{Eigenvector_comp}
\end{figure}

We need a metric to capture the differences in distributions of various datasets compared to the buyer's dataset and reflect it in diversity and relevance.
Our approach focuses on the second moment to capture the variations in distribution. 
In particular, we consider principal component analysis (PCA) to capture the properties of data distributions at different nodes, where it measures the variance of data in directions corresponding to the principal components.
We first find the principal components, together with their corresponding variance values, of the covariance matrix at the buyer.
Then, the principal components of the buyer's data are shared with each seller, and he reports the variance of his covariance matrix in those directions. 
We then use the volume corresponding to the difference and intersection of the variances in the principal components directions to estimate diversity and relevance of the seller's data compared to the buyer's data, respectively. 
This is demonstrated in Fig. \ref{Eigenvector_comp}.

To be precise, the buyer first applies eigendecomposition to the covariance matrix $\frac{1}{n_b} \boldsymbol{B}^T \boldsymbol{B}$, which results in $\frac{1}{n_b} \boldsymbol{B}^T \boldsymbol{B} = \begin{bmatrix} \boldsymbol{u}_1 & \boldsymbol{u}_2 \end{bmatrix} \mbox{Diag} (\lambda_1, \lambda_2) \begin{bmatrix} \boldsymbol{u}_1 & \boldsymbol{u}_2 \end{bmatrix}^T$,
where $\boldsymbol{u}_1 = \begin{bmatrix} 0.99 & 0.13 \end{bmatrix}^T$ and $\boldsymbol{u}_2 = \begin{bmatrix} -0.13 & 0.99\end{bmatrix}^T$ 
are the eigenvectors (principal components), and $\lambda_1 = 1.01$ and $\lambda_2 = 0.23$ are the eigenvalues (measuring the variance in the direction of their corresponding eigenvectors). 
Next, the seller aims to find the variance of his data in both directions $\boldsymbol{u}_1$ and $\boldsymbol{u}_2$, the eigenvectors of buyer's data.
Having vectors $\boldsymbol{u}_1$ and $\boldsymbol{u}_2$ shared with a seller, he estimates the variance of his data in these directions by first computing the covariance matrix $\frac{1}{n_s} \boldsymbol{S}^T \boldsymbol{S}$, then the $l^2$-norm of this matrix projected onto the directions as follows $\hat{\lambda}_1 = \| \frac{1}{n_s} \boldsymbol{S}^T \boldsymbol{S} \boldsymbol{u}_1 \|$, $\hat{\lambda}_2 = \| \frac{1}{n_s} \boldsymbol{S}^T \boldsymbol{S} \boldsymbol{u}_2 \|$.
We note that, if $\boldsymbol{u}_1$ and $\boldsymbol{u}_2$ are the eigenvectors of matrix $\frac{1}{n_s} \boldsymbol{S}^T \boldsymbol{S}$, then $\hat{\lambda}_1$ and $\hat{\lambda}_2$, given above, are the exact eigenvalues of this matrix. 
However, in general $\hat{\lambda}_1$ and $\hat{\lambda}_2$ are not eigenvalues of $\frac{1}{n_s} \boldsymbol{S}^T \boldsymbol{S}$.
Accordingly, at seller 3 with data generated according to $\mathcal{N} (\boldsymbol{0}_{2}, [[0.5, 0.2], [0.2, 0.5]])$, after receiving vectors $\boldsymbol{u}_1$ and $\boldsymbol{u}_2$ from the buyer, we have $\hat{\lambda}_1 = 0.53$ and $\hat{\lambda}_2 = 0.47$.

Fig. \ref{Eigenvector_comp} illustrates vectors $\lambda_1 \boldsymbol{u}_1$ and $\lambda_2 \boldsymbol{u}_2$, the principal components of the buyer's data, as well as $\hat{\lambda}_1 \boldsymbol{u}_1$ and $\hat{\lambda}_2 \boldsymbol{u}_2$, the variance estimate of seller 3's data in the directions of $\boldsymbol{u}_1$ and $\boldsymbol{u}_2$. 
The goal is to estimate diversity and relevance of seller 3's data for the buyer based on the knowledge of $\lambda_1$, $\lambda_2$ and $\hat{\lambda}_1$, $\hat{\lambda}_2$. 
We argue that the volume measuring the difference (shown by red dots) represents diversity, while the intersection volume (shown by green dots) represents the relevance that seller 3's data has for the buyer.
The rationale behind these choices is that the volume capturing the difference ($| \lambda_1 - \hat{\lambda}_1 | \times | \lambda_2 - \hat{\lambda}_2 |$) represents the dissimilarity between the two distributions measured through the second moment on principal components of the buyer's data, which is translated into the diversity of seller 3's data for the buyer.
Whereas, the intersection volume ($\min\{ \lambda_1 , \hat{\lambda}_1 \} \times \min\{ \lambda_2 , \hat{\lambda}_2 \}$) measures the similarity between the two distributions through second moment, which is translated to the relevance of seller 3's data for the buyer.

In order to limit diversity and relevance values to the interval $[0, 1]$, we divide each of the difference and intersection volumes by the whole volume, i.e., $\max\{ \lambda_1 , \hat{\lambda}_1 \} \times \max\{ \lambda_2 , \hat{\lambda}_2 \}$. 
Furthermore, we take the square root of the result to account for the geometric mean; that is, we estimate diversity and relevance, respectively, as follows:
\begin{subequations}
\begin{align}
\mbox{Div.} &= \bigg( \frac{\big| \lambda_1 - \hat{\lambda}_1 \big| \times \big| \lambda_2 - \hat{\lambda}_2 \big|}{\max\{ \lambda_1 , \hat{\lambda}_1 \} \times \max\{ \lambda_2 , \hat{\lambda}_2 \}} \bigg)^{1/2}, \\
\mbox{Rel.} &= \bigg( \frac{\min\{ \lambda_1 , \hat{\lambda}_1 \} \times \min\{ \lambda_2 , \hat{\lambda}_2 \}}{\max\{ \lambda_1 , \hat{\lambda}_1 \} \times \max\{ \lambda_2 , \hat{\lambda}_2 \}} \bigg)^{1/2}.    
\end{align}
\end{subequations}
We will show 
in the Appendix that, with the above estimates of diversity and relevance, we have $\mbox{Div.} + \mbox{Rel.} \le 1$.
With the proposed measures, in general the desired output of $(\mbox{Div.}, \mbox{Rel.})$ pair may be around $(0.5, 0.5)$, i.e., the buyer may desire moderate levels of diversity and relevance jointly instead of sacrificing one for the other, although this depends on the buyer's desire which may vary across the buyers.

\begin{figure}[t!]
\centering
\includegraphics[scale=0.52]{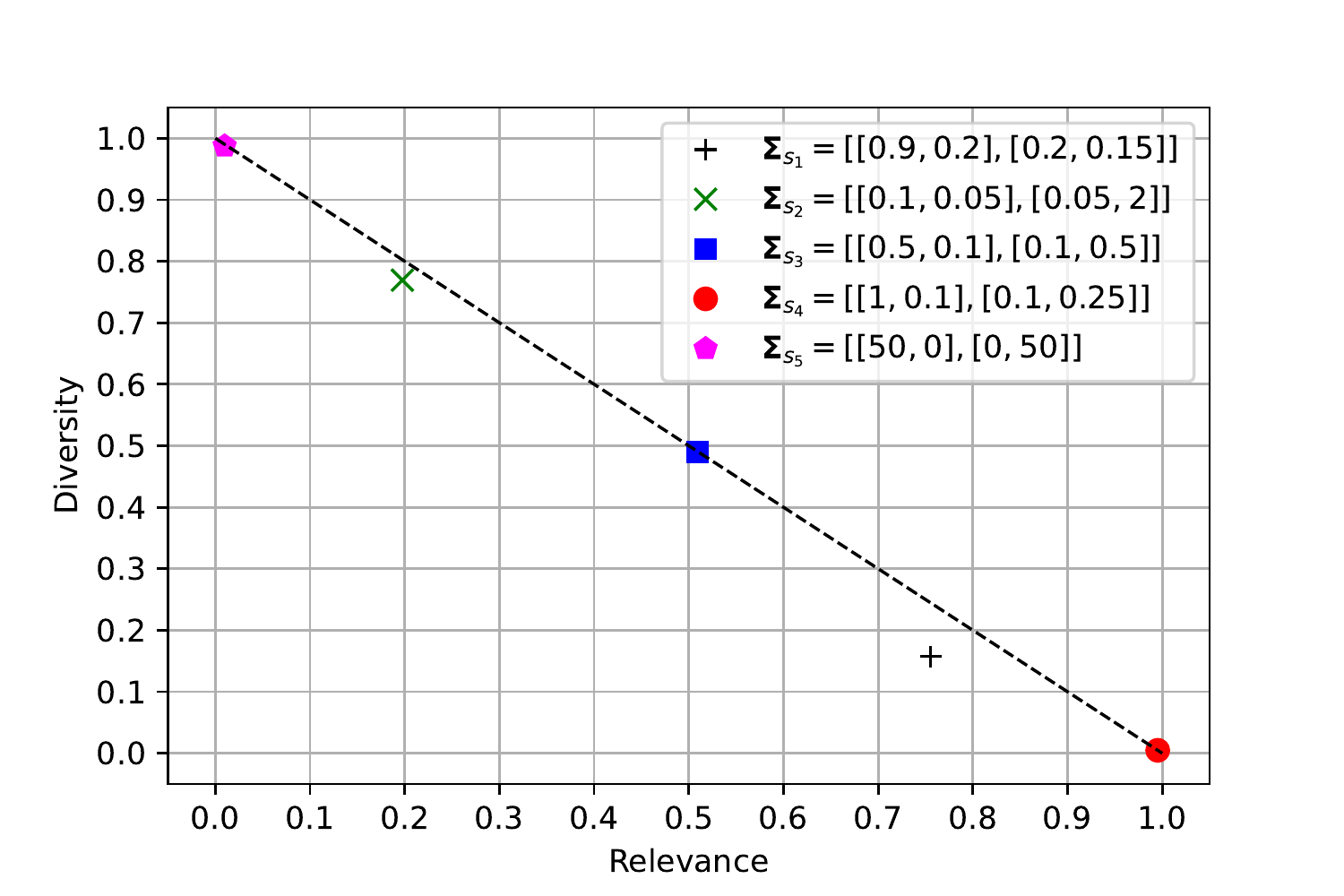}
\caption{Diversity versus relevance of various datasets with 2-D zero-mean Gaussian distributions with various covariance matrices $\boldsymbol{\Sigma}_s$ with respect to the baseline dataset with covariance matrix $\boldsymbol{\Sigma}_b = [[1, 0.1], [0.1, 0.25]]$.}
\label{Fig_div_rel_xmpl}
\end{figure}

Fig. \ref{Fig_div_rel_xmpl} shows the diversity and relevance of various sellers' data for the buyer estimated based on our approach given that the buyer has Gaussian samples with covariance matrix $\boldsymbol{\Sigma}_b = [[1, 0.1], [0.1, 0.25]]$.
As expected intuitively, data at seller 1 with covariance matrix $\boldsymbol{\Sigma}_{s_1} = [[0.9, 0.2], [0.2, 0.15]]$ resembles buyer's data and adds little diversity to it.  
Whereas, seller 2 with data with covariance matrix $\boldsymbol{\Sigma}_{s_2} = [[0.1, 0.05], [0.05, 2]]$ provides the buyer with a diverse data with little relevance. 
Unlike these two sellers, seller 3's data with covariance matrix $\boldsymbol{\Sigma}_{s_3} = [[0.5, 0.1], [0.1, 0.5]]$ has a moderate level of diversity and relevance for buyer with the pair very close to $(0.5, 0.5)$. 
These results indicate that the proposed estimates of diversity and relevance corroborate our intuition.
Also, the proposed approach returns the expected results for the scenarios in Case 1 and Case 2 given the estimated diversity-relevance pair for seller 4's and seller 5's data, respectively.

\vspace{-.0cm}
\section{Diversity and Relevance Estimation}\label{Sec_div_rel}
In this section, we present our approach in estimating diversity and relevance of dataset $\boldsymbol{S} \in \mathbb{R}^{n_s \times d}$ (seller's data) compared to the baseline dataset $\boldsymbol{B} \in \mathbb{R}^{n_b \times d}$ (buyer's data).
This is carried out by comparing the statistical properties of the two datasets through second moment. 
In particular, we compare the variance of data at the seller and buyer in the directions of principal components of buyer's data.

The buyer employs eigendecomposition to the covariance matrix $\frac{1}{n_b} \boldsymbol{B}^T \boldsymbol{B}$; i.e., $\frac{1}{n_b} \boldsymbol{B}^T \boldsymbol{B} = \boldsymbol{U} \; \mbox{Diag} (\lambda_1, ..., \lambda_d) \; \boldsymbol{U}^T$, 
where $\lambda_i$ is the $i$-th largest eigenvalue, and $\boldsymbol{U} = \begin{bmatrix}
\boldsymbol{u}_1 \cdots
\boldsymbol{u}_d
\end{bmatrix}$ with $\boldsymbol{u}_i \in \mathbb{R}^d$ denoting the eigenvector corresponding to the $i$-th eigenvalue.
We note that $\lambda_i \ge 0$ since $\frac{1}{n_b} \boldsymbol{B}^T \boldsymbol{B}$ is positive semi-definite.
The buyer shares the principal components $\boldsymbol{u}_1, ..., \boldsymbol{u}_d$ with the seller through the broker, while $\lambda_1, ..., \lambda_d$ stay local at the buyer. 
The seller aims to estimate the variance of its covariance matrix $\frac{1}{n_s} \boldsymbol{S}^T \boldsymbol{S}$ in the directions of $\boldsymbol{u}_1, ..., \boldsymbol{u}_d$.
This is carried out as follows:
\begin{align}\label{eq_hat_lambdai}
\hat{\lambda}_i = \big\| \frac{1}{n_s} \boldsymbol{S}^T \boldsymbol{S} \boldsymbol{u}_i \big\|, \quad i=1, ..., d,     
\end{align}
where the covariance matrix $\frac{1}{n_s} \boldsymbol{S}^T \boldsymbol{S}$ is first projected into $\boldsymbol{u}_i$ and then $l^2$-norm of the resultant vector provides the estimate of the variance (the data matrices are zero-centered). 
We note that if $\boldsymbol{u}_i$ is an eigenvector of $\frac{1}{n_s} \boldsymbol{S}^T \boldsymbol{S}$, then $\hat{\lambda}_i$ is its corresponding eigenvalue. 
Next, seller and buyer share $\hat{\lambda}_i$ and ${\lambda}_i$, for $i=1, ..., d$, respectively, with the broker.
The broker then tries to estimate the diversity and relevance of the seller's data for the buyer according to $\hat{\lambda}_i$ and ${\lambda}_i$.

\begin{figure*}[t!]
\centering
\includegraphics[scale=0.7]{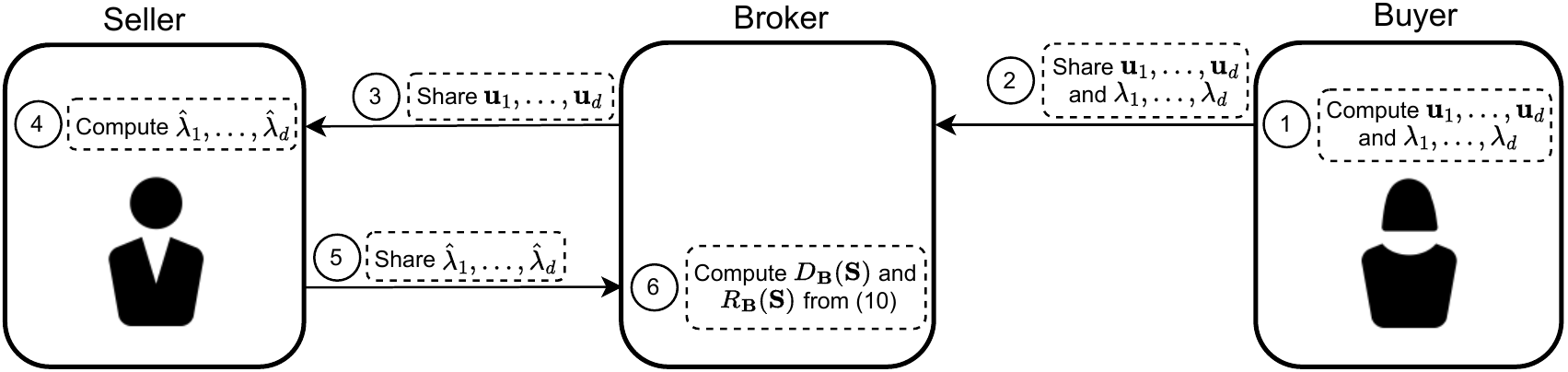}
\caption{The proposed interaction between different parties to estimate diversity and relevance of seller's data for buyer.}
\label{Fig_div_rel_algorithm}
\end{figure*}

We estimate diversity and relevance based on the volume of the space specified by the coordinates corresponding to the principal components (eigenvectors) of the covariance matrix of buyer's data.  
We have $\lambda_i$ and $\hat{\lambda}_i$ as the value of buyer's and seller's data on the $i$-th coordinate, respectively. 
We estimate the relevance through the volume occupied by both buyer's and seller's data in these coordinates; that is, $\prod_{i=1}^{d} \min \{ \lambda_i, \hat{\lambda}_i\}$. 
Our justification is that this volume captures the similarity in the statistical properties of the two datasets since this is a space occupied by both datasets.
On the other hand, diversity is estimated through the volume of the difference between the variance of the buyer's and seller's data in each coordinate; that is, $\prod_{i=1}^{d} | \lambda_i - \hat{\lambda}_i |$. 
We argue that this volume captures the amount of dissimilarity in the statistical properties of the two datasets.  
We normalize these estimates through dividing it by the entire volume, i.e., $\prod_{i=1}^{d} \max \{ \lambda_i, \hat{\lambda}_i\}$.
This yields the following estimates for diversity and relevance, respectively, $\prod_{i=1}^{d} \Big( \frac{| \lambda_i - \hat{\lambda}_i |}{\max\{ \lambda_i , \hat{\lambda}_i \}} \Big)$, $\prod_{i=1}^{d} \Big(\frac{\min \{ \lambda_i, \hat{\lambda}_i\}}{\max\{ \lambda_i , \hat{\lambda}_i \}} \Big)$.
Each of the above estimates is the product of $d$ terms each $\le 1$; so for large enough $d$, these estimates may be very close to $0$. To address this issue, we take the geometric mean, and estimate diversity and relevance of seller's data $\boldsymbol{S}$ for buyer with data $\boldsymbol{B}$, respectively, as follows:
\begin{subequations}\label{eq_div_rel_w_geo_mean}
\begin{align}
D_{\boldsymbol{B}} (\boldsymbol{S}) &= \prod_{i=1}^{d} \bigg( \frac{\big| \lambda_i - \hat{\lambda}_i \big|}{\max\{ \lambda_i , \hat{\lambda}_i \}} \bigg)^{1/d}, \\ R_{\boldsymbol{B}} (\boldsymbol{S}) &= \prod_{i=1}^{d} \bigg(\frac{\min \{ \lambda_i, \hat{\lambda}_i\}}{\max\{ \lambda_i , \hat{\lambda}_i \}} \bigg)^{1/d}.    
\end{align}
\end{subequations}
Fig. \ref{Fig_div_rel_algorithm} shows the proposed approach with the interactions between different parties to value a seller's data for a buyer.

It is easy to verify that the proposed diversity and relevance estimates satisfy the conditions in Case 1 and Case 2. 
We also show in the Appendix that the proposed estimates validate additional intuitive properties.
Generally speaking, with the proposed approach a safe default target is to have a diversity-relevance pair close to $(0.5, 0.5)$. 
However, this may change depending on a buyer's desire. 
For instance for a buyer with a relatively small amount of data, acquiring a highly diverse dataset may not be desirable since it is likely that the sample complexity requirements will not be satisfied given her own limited data samples
(in other words, adding diverse data samples complexifies the underlying distribution, which increases the already excessive sample complexity).
While, a buyer with a relatively large amount of data may prefer acquiring a more diverse data since most likely she already has enough samples to generalize to her own data distribution
(she has enough room to diversify her underlying distribution and increase her sample complexity).    

\noindent \textbf{Partial Components.}
We remark that the proposed approach can be readily extended to the scenario where diversity and relevance could be estimated using the variance in the directions of only partial main components rather than all the $d$ directions; that is, assuming a subset $\mathcal{D} \subset \{1, ..., d\}$,   
\begin{subequations}\label{eq_div_rel_w_geo_mean_partial}
\begin{align}
D_{\boldsymbol{B}} (\boldsymbol{S}) &= \prod_{i\in \mathcal{D}} \bigg( \frac{\big| \lambda_i - \hat{\lambda}_i \big|}{\max\{ \lambda_i , \hat{\lambda}_i \}} \bigg)^{1/| \mathcal{D} |}, \\ R_{\boldsymbol{B}} (\boldsymbol{S}) &= \prod_{i\in \mathcal{D}} \bigg(\frac{\min \{ \lambda_i, \hat{\lambda}_i\}}{\max\{ \lambda_i , \hat{\lambda}_i \}} \bigg)^{1/| \mathcal{D} |}.    
\end{align}
\end{subequations}
This is valid with high dimensional data since not all the principal components carry significant information (\citeauthor{TutorialPrincipalComponentAnalysis}).  

\noindent  \textbf{Representations.}
Assuming that different parties have access to the same publicly available pre-trained model (such as VGG16 trained on the ImageNet dataset (\citeauthor{ImageNetPaepr})), the proposed estimates to measure diversity and relevance can be employed to the representations of data, instead of raw data.
To be precise, different parties can forward propagate the data to the (same) pre-trained model and obtain the activations of the last hidden layer of the model, and then apply the proposed algorithm to estimate diversity and relevance between different datasets.  
We highlight that the last hidden layer output provides a compact representations of data by capturing its most significant attributes (\citeauthor{UnreasonableEffectivenessFeaturesPerceptualMetric}).


\vspace{-.0cm}
\section{Experiments}\label{Sec_experiments}

We evaluate our estimates for diversity and relevance using real datasets, namely Adult (\citeauthor{AdultDataSet}), MNIST (\citeauthor{LeCunMNIST}), fashion-MNIST (\citeauthor{FMNISTDataSet}), Cifar-10 (\citeauthor{Cifar10DatasetKrizhevsky}) and FairFace (\citeauthor{FairFaceDataSet}) datasets.
Throughout the experiments, we estimate diversity and relevance through the partial components analysis, given in \eqref{eq_div_rel_w_geo_mean_partial}, where only the principal components of buyer's data with corresponding eigenvalues more than $10^{-2}$ are chosen.  
We will observe in the experiments that the proposed approach is able to capture the increase in diversity (relevance) when reducing (enhancing) the overlaps in demographics or labels between seller's and buyer's datasets;
It can also capture the introduced diversity due to adding random noise to the seller's data.

\begin{figure}[t!]
\centering
\includegraphics[scale=0.54]{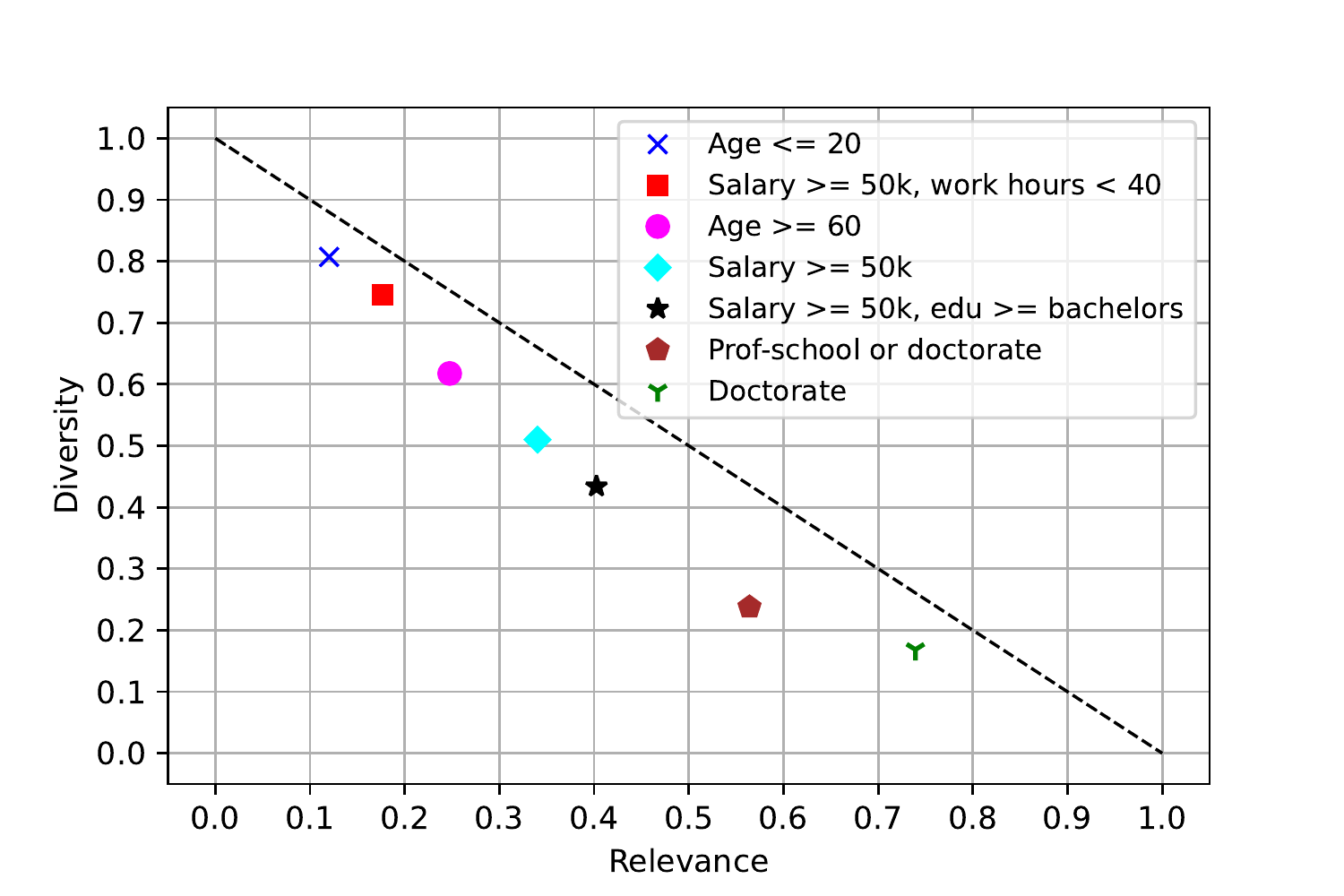}
\caption{Diversity versus relevance of datasets at various sellers compared to buyer's dataset considering the Adult dataset, where the buyer has data of individuals with doctorate degree earning an annual salary of at least 50k.}
\label{Fig_div_rel_Adults}
\end{figure}

We first consider the Adult dataset which has various features such as education level, age, occupation, gender, etc., predicting whether an individual's annual income is over 50k or not. 
We consider a buyer with a dataset of individuals with education level doctorate and annual salary over 50k. 
We aim to measure the amount of diversity and relevance of datasets in various sellers for this buyer. 
We consider a list of sellers with datasets of individuals with: i) no more than 20 years old; ii) annual salary more than 50k and working for less than 40 hours per week; iii) 60 years of age and older; iv) annual salary more than 50k; v) annual salary more than 50k and education level of at least bachelors, i.e., bachelors, masters, prof-school, or doctorate; vi) ducation level of prof-school or doctorate; vii) education level of doctorate.

Fig. \ref{Fig_div_rel_Adults} illustrates the diversity-relevance pair of each seller's data for the buyer using the proposed estimates. 
As expected, we observe that the seller with data only from the individuals not older than 20 years provides the most diverse and least relevant data;
it is highly likely that none of these individuals hold a doctorate degree and earn at least 50k per year. 
Also, it is likely that most of those with doctorate degree and annual salary of at least 50k work for more than 40 hours per week;
that is why the seller with information about individuals working less than 40 hours per week while earning more than 50k annually has more diversity than relevance for the buyer. 
We also expect that most people over the age of 60 years do not hold doctorate degree and/or earn more than 50k per year. 
However, as we expect, for the buyer the relevance of data at the 3rd seller is more than that at the first seller which is also reflected in our estimates.

Considering a seller with data of individuals earning at least 50k annually, this provides a slightly more diversity than relevance for the buyer, since most of these individuals may not hold a doctorate degree;
the diversity however is smaller than the data of individuals older than 60 years which includes people with more diverse education levels. 
Limiting the dataset of high income individuals to those holding bachelors degree or higher academic degree (masters, prof-school, doctorate) reduces the diversity and increases the relevance to the buyer's data.  
On the other hand, the seller with data of individuals having doctorate degree provides a relatively high relevance for the buyer (since most of those probably earn more than 50k annually), while the relevance reduces considering a seller with information of people holding degree from a professional school or doctorate.
We observe that our estimates capture these differences. 

\begin{figure}[t!]
\centering
\includegraphics[scale=0.54]{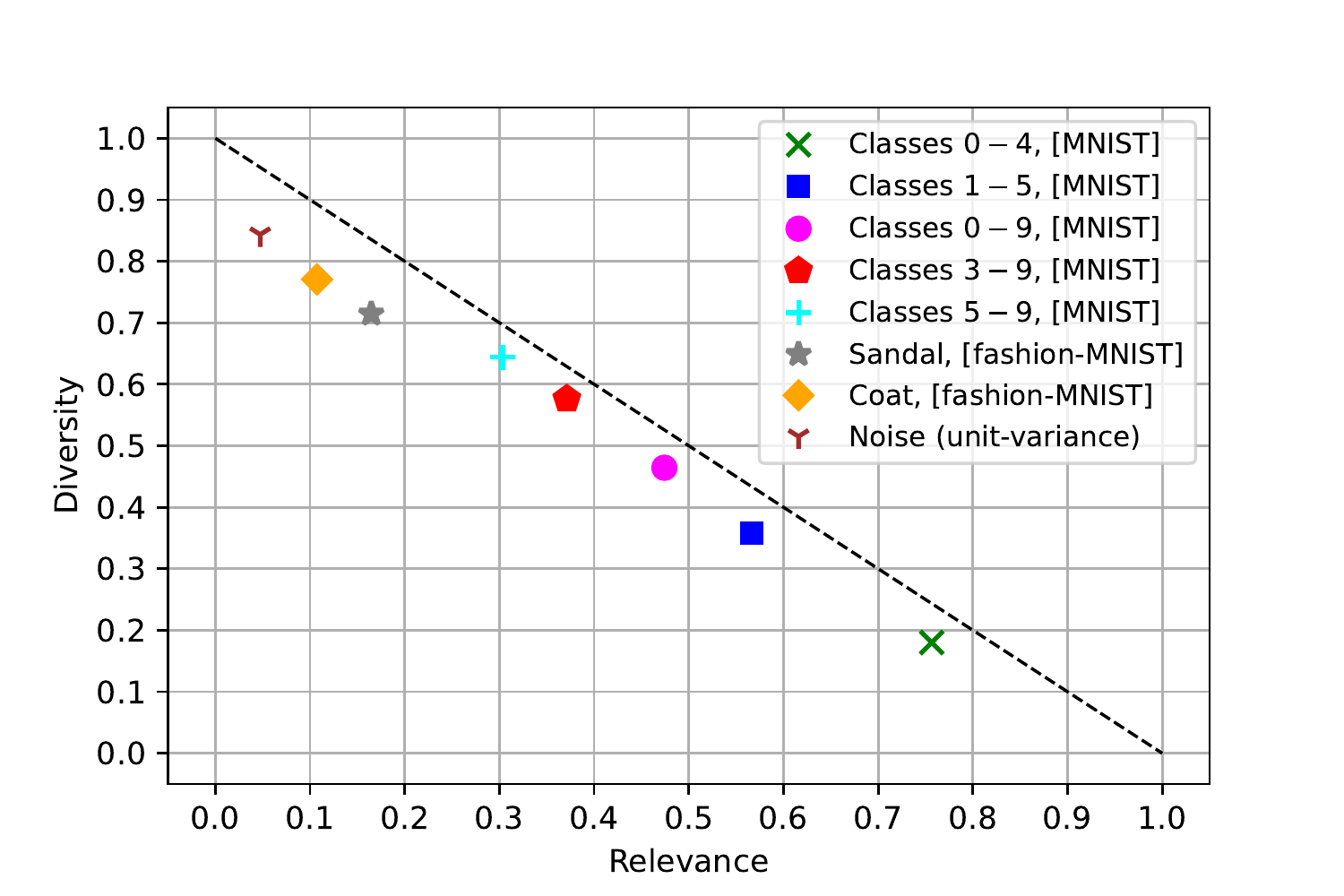}
\caption{Diversity versus relevance of data at various sellers with data from MNIST, fashion-MNIST and noisy images compared to buyer's data from MNIST with classes 0 to 4.}
\label{Fig_div_rel_MNIST_FMNIST}
\end{figure}

Fig. \ref{Fig_div_rel_MNIST_FMNIST} evaluate our estimates of diversity and relevance using MNIST and fashion-MNIST datasets. 
We assume that the buyer has $\sim 6000$ images and sellers have $\sim 10000$ images, and each party has access to distinct images. 
We consider buyer with images of only classes 0 to 4 from MNIST, and five sellers with images from MNIST with different classes, precisely classes 0 to 4 (same as the buyer), 1 to 5, 0 to 9, 3 to 9, and 5 to 9.
It is evident that the diversity/relevance should increase/decrease from seller 1 to seller 5, where the proposed estimates illustrate these changes. 
It is interesting to note that the seller with data from all the classes 0 to 9 provides a diversity-relevance pair close to point $(0.5, 0.5)$. 
To further evaluate the proposed estimates, we consider sellers with images of a different dataset than the images at the buyer. 
In particular, we consider two sellers with images of Sandal and Coat classes, respectively, from fashion-MNIST dataset.
As expected, we observe that these two sellers provide a more diverse data for the buyer compared to the sellers having images from MNIST. 
Furthermore, we consider a seller with only noisy images where each pixel (in a $28 \times 28$ image in this case) is assumed to have a zero-mean unit-variance Gaussian distribution.
This results in a noisy image with a relatively negligible relevance for the buyer; 
this seller is reported by the proposed estimates to have the largest diversity and smallest relevance to the buyer's data compared to the other sellers.
We note that by increasing the noise variance, the diversity-relevance pair will get closer to point $(1, 0)$, i.e., corresponding to Case 2.

\begin{figure}[t!]
\centering
\includegraphics[scale=0.54]{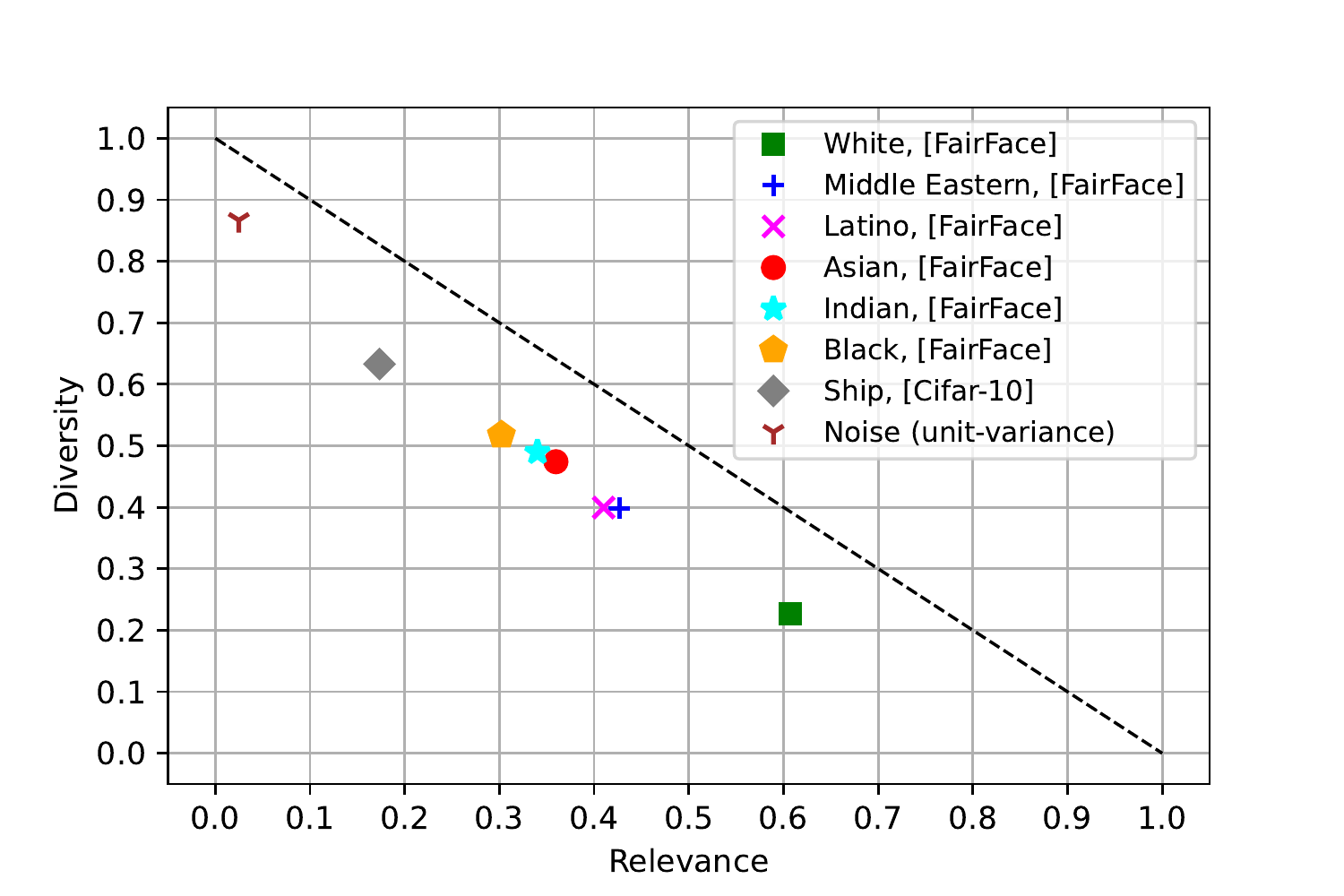}
\caption{Diversity versus relevance of data at various sellers from Cifar-10, FairFace and noisy images compared to buyer's data from FairFace with images of class White.}
\label{Fig_div_rel_FairFace_Cifar10}
\end{figure}

Next we consider a buyer with images from the FairFace dataset, which contains face images of different race groups, namely White, Black, Indian, Middle Eastern, Asian (combining East Asian and Southeast Asian).
We assume that the buyer and all the sellers have access to the publicly available VGG16 model pre-trained on the ImageNet dataset, and apply the proposed scheme to estimate diversity and relevance to the output of the last hidden layer of this pre-trained model (feature representations of data).
We assume that the buyer has images of only White group, and six sellers each has images of one group (the images of White group at a seller are different than the ones at the buyer).
We observe in Fig. \ref{Fig_div_rel_FairFace_Cifar10} that the images of Black group provides the most diversity for the buyer, and images of Indian and Asian (East Asian + Southeast Asian) have respectively more diverse than relevant data for the buyer. 
Whereas, sellers with images of Middle Eastern and Latino groups are more relevant for the buyer compared to the other three groups, and, as expected, the seller with White group images has the least diverse data for the buyer.
Overall, human face images follow a particular pattern and does not vary significantly across individuals. 
To further capture the differences, we consider a seller with Ship images from Cifar-10 passed through the pre-trained VGG16 model on ImageNet. 
We observe that this seller provides a relatively small relevant data for the buyer. 
Also, we consider a seller with only noisy images, where the proposed estimates report this seller as one with highly diverse and negligible relevant data for the buyer.

\vspace{-.0cm}
\section{Discussions}\label{Sec_diss}
Here we discuss about various aspects of the proposed approach ranging from its properties to possible extensions.


\noindent \textbf{Privacy Enhancing.}
With the proposed approach, the only information about buyer's data that is shared with the seller is all/partial directions of the principal components of its covariance matrix.
Our approach could be easily modified to enhance the privacy of buyer's data through sharing extra directions, in addition to that of the principal components, with the seller.
This can hide the principal components of buyer's data, which are the only directions used to estimate diversity and relevance at the broker. 

\noindent \textbf{Robustness to Malicious Seller(s).} 
With the proposed approach, the seller does not have any information about the variance of buyer's data in different directions.
This provides a robust mechanism to malicious sellers who try to fabricate their data resulting in a particular diversity-relevance pair and/or add noise to their data;
that is, without the knowledge of the variances of buyer's data in different directions, it is impossible to manipulate the algorithm to output a specific diversity-relevance pair.
Also, adding random noise to the data at the seller may increase the diversity for the buyer, however it reduces the relevance.
Thus, the trade-off between diversity and relevance controls a malicious seller in adding noise to his data for more diversity. 

\noindent \textbf{Number of Data Samples.} 
We assume that sellers have large enough data compared to the buyer, and we do not incorporate the size of data at the sellers into the proposed data valuation formulation.
This can be extended by considering the size of data at the sellers as an additional metric for data valuation.
We can further extend the approach by considering the mean of data samples at different parties.

\noindent \textbf{Weighted Averaging.}
The proposed diversity and relevance estimates could be extended by weighting the ratios at various principal components of the buyer's data differently. 
For weights $\omega_1, ..., \omega_d$ with $0 \le \omega_i \le 1$, we can estimate diversity as $D_{\boldsymbol{B}} (\boldsymbol{S}) = \prod_{i=1}^{d} \Big(\omega_i \frac{\left| \lambda_i - \hat{\lambda}_i \right|}{\max\{ \lambda_i , \hat{\lambda}_i \}} \Big)^{1/d}$ and relevance and $R_{\boldsymbol{B}} (\boldsymbol{S}) = \prod_{i=1}^{d} \left(\omega_i \frac{\min \{ \lambda_i, \hat{\lambda}_i\}}{\max\{ \lambda_i , \hat{\lambda}_i \}} \right)^{1/d}$.
This is useful when the variance in one particular direction may be of more importance for the buyer. 
Furthermore, 
we can output a single value as the data value computed as the combination of diversity and relevance specified by the buyer.
For example, if the buyer prefers to have ratio $\alpha$ diverse and $1-\alpha$ relevant data, for $0 \le \alpha \le 1$, data value of a seller for the buyer can be computed as $\alpha D_{\boldsymbol{B}} (\boldsymbol{S}) + (1 - \alpha) R_{\boldsymbol{B}} (\boldsymbol{S})$.

\noindent \textbf{Maximum Diversity Maximum Relevance.}
With the proposed approach, the diversity and relevance estimates are not independent since their sum can not exceed 1.
Although ideally maximum diversity 1 and maximum relevance 1 may be desired, we argue that this may not be feasible. 
Maximum diversity encompasses all the randomness that could be added to the data for more diversity; that is, it is a case where adding more randomness does not increase the diversity. 
Having maximum relevance in such a case may not be feasible. 
Thus, as derived by the proposed approach, we believe that there is an inverse relationship between diversity and relevance reflected to their sum being limited.

\vspace{-.05cm}
\section{Conclusions}\label{SecConclucsions}
We studied task-agnostic valuation of data at a seller for a buyer.  
This is specifically relevant when the buyer has access to data samples apriori which could be used to measure usefulness of seller's data for the buyer.
We formulated the problem as the diversity and relevance of the seller's data for the buyer in the efforts to compare the statistical properties of the two datasets. 
We then provided estimates for the diversity and relevance metrics by measuring the difference and similarity volumes using the space of the principal components of the buyer's data as the baseline;
this technique focuses on the second moment analysis and is carried out through comparing the variance of each dataset on these components.
We show that the proposed estimates are successful in capturing the diversity and relevance of two datasets using various real datasets.












\bibliography{aaai23}



\appendix

\section{Appendix}

\subsection{Diversity and Relevance Relationship}
We show that, the diversity and relevance estimates given in \eqref{eq_div_rel_w_geo_mean} results in $D_{\boldsymbol{B}} (\boldsymbol{S}) + R_{\boldsymbol{B}} (\boldsymbol{S}) \le 1$.
We have
\newcommand\ainequal{\mathrel{\overset{\makebox[0pt]{\mbox{\normalfont\tiny\sffamily (a)}}}{\le}}}
\newcommand\bequal{\mathrel{\overset{\makebox[0pt]{\mbox{\normalfont\tiny\sffamily (b)}}}{=}}}
\begin{align}\label{Div_plus_rel_ineq}
& D_{\boldsymbol{B}} (\boldsymbol{S}) + R_{\boldsymbol{B}} (\boldsymbol{S}) \nonumber\\
& = \prod_{i=1}^{d} \bigg( \frac{\big| \lambda_i - \hat{\lambda}_i \big|}{\max\{ \lambda_i , \hat{\lambda}_i \}} \bigg)^{1/d} + \prod_{i=1}^{d} \bigg(\frac{\min \{ \lambda_i, \hat{\lambda}_i\}}{\max\{ \lambda_i , \hat{\lambda}_i \}} \bigg)^{1/d} \nonumber\\
& \ainequal \frac{1}{d} \sum_{i=1}^{d} \bigg( \frac{\big| \lambda_i - \hat{\lambda}_i \big|}{\max\{ \lambda_i , \hat{\lambda}_i \}} \bigg) + \frac{1}{d} \sum_{i=1}^{d} \bigg(\frac{\min \{ \lambda_i, \hat{\lambda}_i\}}{\max\{ \lambda_i , \hat{\lambda}_i \}} \bigg) \nonumber\\
& = \frac{1}{d} \sum_{i=1}^{d} \bigg( \frac{\big| \lambda_i - \hat{\lambda}_i \big| + \min \{ \lambda_i, \hat{\lambda}_i\}}{\max\{ \lambda_i , \hat{\lambda}_i \}} \bigg) \bequal 1,
\end{align}
where (a) follows from inequality $\prod_{i=1}^{d} (a_i)^{1/d} \le \frac{1}{d} \sum_{i=1}^{d} a_i$, for positive values $a_1, ..., a_d$,
and (b) follows since $\big| \lambda_i - \hat{\lambda}_i \big| + \min \{ \lambda_i, \hat{\lambda}_i\} = \max\{ \lambda_i , \hat{\lambda}_i \}$ for $\lambda_i \ge 0$.

\subsection{Properties of Our Proposed Estimates}
We provide some interesting properties of the proposed approach to estimate diversity and relevance showing that it corroborates our intuition. 
Let $\boldsymbol{u}_i$ and $\lambda_i$ denote eigenvector and eigenvalue of $\frac{1}{n_b} \boldsymbol{B}^T \boldsymbol{B}$ at the buyer, respectively.
We consider two different sellers, seller 1 and seller 2, with data matrices $\boldsymbol{S}_1$ and $\boldsymbol{S}_2$ with variances $\lambda_{s_1,i}$ and $\lambda_{s_2,i}$ in the direction of $\boldsymbol{u}_i$, respectively, computed according to \eqref{eq_hat_lambdai}.
The diversity and relevance of the sellers for the buyer is estimated according to \eqref{eq_div_rel_w_geo_mean}.
We assume that $\lambda_{s_1,i}$ and $\lambda_{s_2,i}$ are the same for all $i$ except for $i=j$; that is, $\lambda_{s_1,i} = \lambda_{s_2,i}$, $\forall i \in \{1, ..., d\} / \{ j \}$.  
We show that the proposed approach in estimating diversity and relevance satisfies the following three intuitive properties:
\begin{itemize}
\item \textbf{Property 1:} Given $\lambda_{s_1,j} \ge \lambda_{j}$, if $\lambda_{s_2,j} \ge \lambda_{s_1,j}$, then it is expected that seller 2's data is more diverse and less relevant for the buyer compared to seller 1's data, and vice versa if $\lambda_{s_1,j} \ge \lambda_{s_2,j} \ge \lambda_{j}$; 
that is, getting further far from the variance of the buyer's data by increasing the variance in one particular direction should lead to more diversity and less relevance.

\textbf{Proof:} For the diversity estimate, we have
\begin{align}\label{Appx_proof_large_lambdas_div}
\frac{D_{\boldsymbol{B}} (\boldsymbol{S}_2)}{D_{\boldsymbol{B}} (\boldsymbol{S}_1)} &= \bigg( \frac{ \lambda_{s_2, j} - {\lambda}_j }{\lambda_{s_2, j}} \bigg)^{1/d} \bigg( \frac{ \lambda_{s_1, j} }{\lambda_{s_1, j} - {\lambda}_j} \bigg)^{1/d} \nonumber\\
& = \bigg(\frac{\lambda_{s_1, j} \lambda_{s_2, j} - \lambda_{j} \lambda_{s_1, j}}{\lambda_{s_1, j} \lambda_{s_2, j} - \lambda_{j} \lambda_{s_2, j}}\bigg)^{1/d}\nonumber\\
& \begin{cases}
\ge 1, & \mbox{if } \lambda_{s_2, j} \ge \lambda_{s_1, j},\\
\le 1, & \mbox{if } \lambda_{s_2, j} \le \lambda_{s_1, j}.
\end{cases}
\end{align}
This proves this property for the diversity estimate. 
Now we consider the relevance estimate
\begin{align}\label{Appx_proof_large_lambdas_rel}
\frac{R_{\boldsymbol{B}} (\boldsymbol{S}_2)}{R_{\boldsymbol{B}} (\boldsymbol{S}_1)} &= \bigg( \frac{{\lambda}_j }{\lambda_{s_2, j}} \bigg)^{1/d} \bigg( \frac{ \lambda_{s_1, j} }{{\lambda}_j} \bigg)^{1/d}\nonumber\\
& \begin{cases}
\ge 1, & \mbox{if } \lambda_{s_2, j} \le \lambda_{s_1, j},\\
\le 1, & \mbox{if } \lambda_{s_2, j} \ge \lambda_{s_1, j},
\end{cases}
\end{align}
which proves the property for the relevance estimate.

\item \textbf{Property 2:} Given $\lambda_{s_1,j} \le \lambda_{j}$, if $\lambda_{s_2,j} \le \lambda_{s_1,j}$, then it is expected that seller 2's data is less diverse and more relevant for the buyer compared to seller 1's data, and vice versa if $\lambda_{s_1,j} \le \lambda_{s_2,j} \le \lambda_{j}$; 
that is, getting closer to the variance of the buyer's data by increasing the variance in one particular direction should lead to less diversity and more relevance.

\textbf{Proof:} For the diversity estimate, we have
\begin{align}\label{Appx_proof_small_lambdas_div}
\frac{D_{\boldsymbol{B}} (\boldsymbol{S}_2)}{D_{\boldsymbol{B}} (\boldsymbol{S}_1)} &= \bigg( \frac{ {\lambda}_j - \lambda_{s_2, j} }{\lambda_{j}} \bigg)^{1/d} \bigg( \frac{ \lambda_{j} }{{\lambda}_j - \lambda_{s_1, j}} \bigg)^{1/d} \nonumber\\
& \begin{cases}
\ge 1, & \mbox{if } \lambda_{s_2, j} \le \lambda_{s_1, j},\\
\le 1, & \mbox{if } \lambda_{s_2, j} \ge \lambda_{s_1, j}.
\end{cases}
\end{align}
This proves this property for the diversity estimate. 
Now we consider the relevance estimate
\begin{align}\label{Appx_proof_small_lambdas_rel}
\frac{R_{\boldsymbol{B}} (\boldsymbol{S}_2)}{R_{\boldsymbol{B}} (\boldsymbol{S}_1)} &= \bigg( \frac{{\lambda}_{s_2, j} }{\lambda_{j}} \bigg)^{1/d} \bigg( \frac{ \lambda_{j} }{{\lambda}_{s_1, j}} \bigg)^{1/d}\nonumber\\
& \begin{cases}
\ge 1, & \mbox{if } \lambda_{s_2, j} \ge \lambda_{s_1, j},\\
\le 1, & \mbox{if } \lambda_{s_2, j} \le \lambda_{s_1, j},
\end{cases}
\end{align}
which proves the property for the relevance estimate.

\item \textbf{Property 3:} Given $\lambda_{s_1,j} \le \lambda_{j}$ and $\lambda_{s_2,j} \ge \lambda_{j}$, if $\lambda_{s_2,j}/\lambda_{j} \ge \lambda_{j} / \lambda_{s_1, j}$ then it is expected that seller 2's data is more diverse and less relevant for the buyer compared to seller 1's data, and vice versa if $\lambda_{s_2,j}/\lambda_{j} \le \lambda_{j} / \lambda_{s_1, j}$; 
that is, increasing the variance ratio between the seller and buyer in one particular direction should lead to more diversity and less relevance.

\textbf{Proof:} For the diversity estimate, we have
\begin{align}\label{Appx_proof_small_ratio_lambdas_div}
\frac{D_{\boldsymbol{B}} (\boldsymbol{S}_2)}{D_{\boldsymbol{B}} (\boldsymbol{S}_1)} &= \bigg( \frac{ \lambda_{s_2, j} - {\lambda}_j }{\lambda_{s_2, j}} \bigg)^{1/d} \bigg( \frac{ \lambda_{j} }{{\lambda}_j - \lambda_{s_1, j}} \bigg)^{1/d}  \nonumber\\
& = \bigg( \frac{ \lambda_{s_2, j} / {\lambda}_j -1 }{\lambda_{s_2, j}/{\lambda}_j} \bigg)^{1/d} \bigg( \frac{ \lambda_{j}/\lambda_{s_1, j} }{{\lambda}_j / \lambda_{s_1, j} - 1} \bigg)^{1/d} \nonumber\\
& \begin{cases}
\ge 1, & \mbox{if } \lambda_{s_2,j}/\lambda_{j} \ge \lambda_{j} / \lambda_{s_1, j},\\
\le 1, & \mbox{if } \lambda_{s_2,j}/\lambda_{j} \le \lambda_{j} / \lambda_{s_1, j}.
\end{cases}
\end{align}
This proves this property for the diversity estimate. 
Now we consider the relevance estimate
\begin{align}\label{Appx_proof_small_ratio_lambdas_rel}
\frac{R_{\boldsymbol{B}} (\boldsymbol{S}_2)}{R_{\boldsymbol{B}} (\boldsymbol{S}_1)} &= \bigg( \frac{{\lambda}_{j} }{\lambda_{s_2, j}} \bigg)^{1/d} \bigg( \frac{ \lambda_{j} }{{\lambda}_{s_1, j}} \bigg)^{1/d}\nonumber\\
& \begin{cases}
\ge 1, & \mbox{if } \lambda_{s_2,j}/\lambda_{j} \le \lambda_{j} / \lambda_{s_1, j},\\
\le 1, & \mbox{if } \lambda_{s_2,j}/\lambda_{j} \ge \lambda_{j} / \lambda_{s_1, j},
\end{cases}
\end{align}
which proves the property for the relevance estimate.

\item \textbf{Property 4:} Given $\lambda_{s_1,j} \ge \lambda_{j}$ and $\lambda_{s_2,j} \le \lambda_{j}$, if $\lambda_{j} / \lambda_{s_2,j} \ge \lambda_{s_1, j} / \lambda_{j}$ then it is expected that seller 2's data is more diverse and less relevant for the buyer compared to seller 1's data, and vice versa if $\lambda_{j} / \lambda_{s_2,j} \le \lambda_{s_1, j} / \lambda_{j}$; 
that is, decreasing the variance ratio between the seller and buyer in one particular direction should lead to more diversity and less relevance.

\textbf{Proof:} For the diversity estimate, we have
\begin{align}\label{Appx_proof_large_ratio_lambdas_div}
\frac{D_{\boldsymbol{B}} (\boldsymbol{S}_2)}{D_{\boldsymbol{B}} (\boldsymbol{S}_1)} &= \bigg( \frac{ {\lambda}_j - \lambda_{s_2, j} }{\lambda_{j}} \bigg)^{1/d} \bigg( \frac{ \lambda_{s_1,j} }{ \lambda_{s_1, j} - {\lambda}_j} \bigg)^{1/d}  \nonumber\\
& = \bigg( \frac{ \lambda_{j} / {\lambda}_{s_2, j} -1 }{\lambda_j/{\lambda}_{s_2, j}} \bigg)^{1/d} \bigg( \frac{ \lambda_{s_1, j}/\lambda_{j} }{{\lambda}_{s_1, j} / \lambda_j - 1} \bigg)^{1/d} \nonumber\\
& \begin{cases}
\ge 1, & \mbox{if } \lambda_{j}/\lambda_{s_2,j} \ge \lambda_{s_1, j} / \lambda_{j},\\
\le 1, & \mbox{if } \lambda_{j}/\lambda_{s_2,j} \le \lambda_{s_1, j} / \lambda_{j}.
\end{cases}
\end{align}
This proves this property for the diversity estimate. 
Now we consider the relevance estimate
\begin{align}\label{Appx_proof_large_ratio_lambdas_rel}
\frac{R_{\boldsymbol{B}} (\boldsymbol{S}_2)}{R_{\boldsymbol{B}} (\boldsymbol{S}_1)} &= \bigg( \frac{{\lambda}_{s_2, j} }{\lambda_{j}} \bigg)^{1/d} \bigg( \frac{ \lambda_{s_1, j} }{{\lambda}_{j}} \bigg)^{1/d}\nonumber\\
& \begin{cases}
\ge 1, & \mbox{if } \lambda_{j} / \lambda_{s_2,j} \le \lambda_{s_1, j} / \lambda_{j},\\
\le 1, & \mbox{if } \lambda_{j} / \lambda_{s_2,j} \ge \lambda_{s_1, j} / \lambda_{j},
\end{cases}
\end{align}
which proves the property for the relevance estimate.

\end{itemize}

 
\end{document}